\documentclass{article}

\usepackage{PRIMEarxiv}

\usepackage[utf8]{inputenc} % allow utf-8 input
\usepackage[T1]{fontenc}    % use 8-bit T1 fonts
\usepackage{hyperref}       % hyperlinks
\usepackage{url}            % simple URL typesetting
\usepackage{booktabs}       % professional-quality tables
\usepackage{amsfonts}       % blackboard math symbols
\usepackage{nicefrac}       % compact symbols for 1/2, etc.
\usepackage{microtype}      % microtypography
\usepackage{lipsum}
\usepackage{fancyhdr}       % header
\usepackage{graphicx}       % graphics
\usepackage{indentfirst}
\usepackage{caption}
\usepackage{float} 

\hypersetup{
    colorlinks=false,
    pdfborder={0 0 0},
}

\graphicspath{{media/}}

\title{A CNN-LSTM Combination Network for Cataract Detection using Eye Fundus Images
}

\author{
  Dishant Padalia \\
  Department of Electronics and Telecommunications \\
  K.J. Somaiya College of Engineering \\
  Mumbai, India\\
  \texttt{dishant.padalia@somaiya.edu} \\
   \And
  Abhishek Mazumdar \\
  Department of Electronics and Telecommunications \\
  K.J. Somaiya College of Engineering \\
  Mumbai, India\\
  \texttt{a.mazumdar@somaiya.edu} \\
  \And
  Bharati Singh \\
  Department of Electronics and Telecommunications \\
  K.J. Somaiya College of Engineering \\
  Mumbai, India\\
  \texttt{bhartisingh@somaiya.edu} \\
}

\begin{document}
\maketitle

\begin{abstract}
According to multiple authoritative authorities, including the World Health Organization, vision-related impairments and disorders are becoming a significant issue. According to a recent report, one of the leading causes of irreversible blindness in persons over the age of 50 is delayed cataract treatment. A cataract is a cloudy spot in the eye's lens that causes visual loss. Cataracts often develop slowly and consequently result in difficulty in driving, reading, and even recognizing faces.  This necessitates the development of rapid and dependable diagnosis and treatment solutions for ocular illnesses. Previously, such visual illness diagnosis were done manually, which was time-consuming and prone to human mistake. However, as technology advances, automated, computer-based methods that decrease both time and human labor while producing trustworthy results are now accessible. In this study, we developed a CNN-LSTM-based model architecture with the goal of creating a low-cost diagnostic system that can classify normal and cataractous cases of ocular disease from fundus images. The proposed model was trained on the publicly available ODIR dataset, which included fundus images of patients' left and right eyes. The suggested architecture outperformed previous systems with a state-of-the-art 97.53\% accuracy.
\end{abstract}

% keywords can be removed

\section{Introduction}
The human body has been identified as having five major senses. Vision is one of the most integral senses and contributes up to 80\% of brain processing with respect to environmental perception. This can validate the disruption and inconvenience that any form of vision impairment can cause. Cataract is one of the leading to vision impairments and in extreme cases even permanent blindness. As per the World Health Organisation \cite{WHO2022Vision}, approximately 94 million people had cataracts in 2021. Another survey led by the World Health Organisation \cite{S2008Current} suggested that 47.8\% of worldwide blindness, and 51\% of blindness in South Asia, including India is due to cataract.  Cataract refers to the condition where there is a development of a cloudy area on the lens associated with aging and degradation of the eye tissues leading to blurry vision. Cataract can be categorized into nuclear, cortical, posterior subcapsular, and congenital cataracts. 
\\\\
Nuclear cataracts \cite{Clinic2022Cataracts} mainly affect the center of the lens, at first it causes nearsightedness or temporary impairment of vision but with time, the lens turns densely yellow and sometimes brown which slowly impedes the vision to a great extent and leads to difficulty in identifying colors. Cortical Cataracts affect the edges of the lens, at start it creates white wedge-shaped opacities on the outer edge of the cortex and slowly progresses towards the center of the lens. Posterior subcapsular cataracts affect the back of the lens and create a tiny opaque area near the back of the lens which interferes with the person’s reading ability. Lastly, Congenital cataracts are genetic types of cataracts and people are usually born with this condition. Throughout the years, researchers have proposed and developed several Comupter Aided Detection systems for diagnosis of ocular diseases. For an instance, Color Fundus Photography (CFPs) and Optical Coherence Tomography have been used to extract and represent global and local features of the eye for diagnostic purposes. These image results have also been combined with personal demographic data for refined diagnostic results. 
\\\\
In recent years, a plethora of Deep Learning based Neural Network models have been explored and developed by several researchers to automate and enhance the precision of Ocular diagnostics. These models have been showing great promise in ocular detection and classification. This has resulted in an extensive study of Convolutional Neural Networks \cite{o2015introduction} in this space. The models have proved to be successful in segmenting retinal vessels and classifying a specific ocular disease. However, only a few of the existing studies have addressed the task of classifying multiple ocular diseases from fundus images. 
\\\\
In this study, we propose a combination of CNN and LSTM model that would assist the  ophthalmologists to classify fundus images into normal and cataract, rapidly and with high precision. In the proposed system, the input image is first passed to CNN \cite{o2015introduction} layers that extract relevant features from the image and then these features are forwarded to the LSTM \cite{greff2016lstm} layers that, because of its ability to learn long-term dependencies, serves as a classifier.

\section{Related Work}
Image understanding systems that exploit machine learning (ML) techniques have been rapidly evolving in recent years. Convolutional neural networks are becoming a mainstream solution for analyzing medical images \cite{yamashita2018convolutional}. CNN are state-of-the-art applications that extract visual information from a given set of input to carry out generative and descriptive tasks. CNN provides an amalgamation of various techniques that can be exploited for learning image representation and feature classification. The scarcity of experts, expensive consultation charges, and the complexities and anomalies in the medical images are complicated to identify. Therefore, computer-aided deep learning tools can significantly impact diagnosis by providing more accurate results than humans.
\\\\
In this day and age, Ocular diseases, such as diabetic retinopathy, cataract, and age-related macular degeneration, are common and can be detrimental if not treated appropriately. They can be associated with an increased risk of ischemic heart disease death in diabetic patients. Early detection is vital but difficult due to the lack of symptoms’ visibility in the early stages. In ophthalmology, colour fundus photography is an economical and effective tool for early-stage ocular disease screening. C. Li et al. \cite{li2020dense} proposed a Dense Correlation Network (DCNet) model on the backbone of CNN (Resnet - 18, 34, 50, 101, with parameters initialized from ImageNet pretraining, the initial learning rate of 0.007, power of 0.9, an epoch of 50 using binary cross-entropy as loss function) was proposed for the extraction of feature representations and a Spatial Correlation Module (SCM) to exploit correlations. The ODIR dataset \cite{ODIR} (structured real-life ophthalmic dataset of paired fundus images from 5000 patients collected by Shanggong Medical Technology Co., Ltd. from different hospitals/medical centres in China) was used. The final results of the model for Resnet 18,34,50, 101 using SCM are 78.5\%, 80.8\%, 82.2\% and 82.7\% respectively. The Model showed better results with SCM than without SCM. Further accuracy enhancements can be studied using other transfer learning adaptations and custom models.
\\\\
One of the most challenging tasks for ophthalmologists is early screening and diagnosing ocular diseases from fundus images. That is why a computer-aided automated ocular disease detection system is required for the early detection of various ocular disorders using fundus images. N. Dipu et al. \cite{dipu2021ocular} presented a study of four deep learning-based models for targeted ocular disease detection, namely Resnet-34, EfficientNet, MobileNetV2, and VGG-16, on the ODIR dataset consisting of 5000 fundus images that belong to 8 different classes. Each of these classes represents another ocular disease. The VGG-16 model achieved an accuracy of 97.23\%; the Resnet-34 model reached 90.85\%; the MobileNetV2 model provided an accuracy of 94.32\%, and the EfficientNet classification model achieved an accuracy of 93.82\%. Thus out of these models, VGG-16 provided the best accuracy of 97.23\% in classifying ocular diseases from the fundus images.
\\\\
India has a blind population of approximately 15 million, and the sad reality is that 75\% of these cases are curable. The doctor-patient ratio in India is 1:10,000. Studies have found that Diabetic Retinopathy(DR) and Glaucoma are the leading causes of blindness in India. DK. Prasad et al. \cite{prasad2015early} proposed a deep neural network model that helps to detect the presence of diabetic retinopathy and glaucoma at its early stages. It can alert the patients to consult an ophthalmologist from a screening standpoint. The developed model is less complicated and resulted in an accuracy of 80\%. The system will produce accurate results, which is achieved by using a less complex 5-layer network. But more accuracy can be obtained by adding dropout layers and dense layers along with the conv2D layers. These layers improve the system's capability to reduce overfitting and improve its feature extraction property. The dataset used was ODIR. The accuracy obtained was 80\%.
\\\\
Research has it that many leading causes of vision impairment (such as Glaucoma) cannot be cured. Data has shown that, due to the lack of visual symptoms in glaucoma and the shortage of clinical resources over  90\% of glaucoma cases remain undetected.  Several computer aided detection (CAD) systems for many ocular diseases have been proposed  for several leading causes of vision impairment. Retinal fundus or OCT images are used to extract global or local visual features. But different ocular diseases have different characteristics and most existing methods today are specifically designed for single disease detection, and methodology of detection for other diseases is still unclear. Y. Xu et al. \cite{xu2018ocular} proposed a novel model developed based on a unified MKL framework called MKLclm to automatically detect ocular diseases by the effective fusion of personal demographic data, genome information and visual information from retinal fundus images through the incorporation of prelearned SVM classifiers. The data used here is the Singapore Malay Eye Study (SiMES) database from a population-based study (among the 2258 subjects after quality control, there are 100 with glaucoma,122 with Age-related macular disease and 58 with Pathological Myopia). The Model showed an 85.3\% accuracy for Glaucoma prediction, 73.2\% for Age-related macular disease prediction and 88.2\% for Pathological Myopia. These results were better as compared to Single Kernel SVM and MKL. The Model although provides a centralized method of multi disease detection shows low accuracy. Hence, alternate models can be tried to enhance the accuracy.

\section{Methodology}
This section covers the dataset description, data preprocessing and the proposed architecture. The proposed architecture was designed by a combination of CNN (Convolutional Neural Networks) and LSTM (Long Short Term Memory) layers. The CNN layers function as feature extractors, understanding image characteristics, whilst the LSTM layers function as classifiers.

\subsection{Dataset Description}
To train and validate the model performance we used the ODIR dataset for this study. Ocular Disease Intelligent Recognition (ODIR) \cite{ODIR} is a real case based structured ophthalmic database of 5000 patients collected by Shanggong Medical Technology Co., Ltd. from different hospitals/medical centers in China. The dataset contains details of patients’ age, color fundus photographs from left and right eyes and doctors' diagnostic keywords from doctors. The fundus images are captured by various cameras in the market, such as Canon, Zeiss and Kowa, resulting into varied image resolutions. Each patient is classified into at least one of the following eight categories, normal (1140 cases), diabetes (1128 cases), glaucoma (215 cases), cataract (212 cases), AMD (164 cases), hypertension (103 cases), myopia (174 cases), and other diseases/abnormalities (979 cases) based on both CFPs and additional clinical features by trained human readers.

\begin{figure}[H]
  \centering
  \captionsetup{justification=centering,margin=2cm}
  \includegraphics[width=\textwidth]{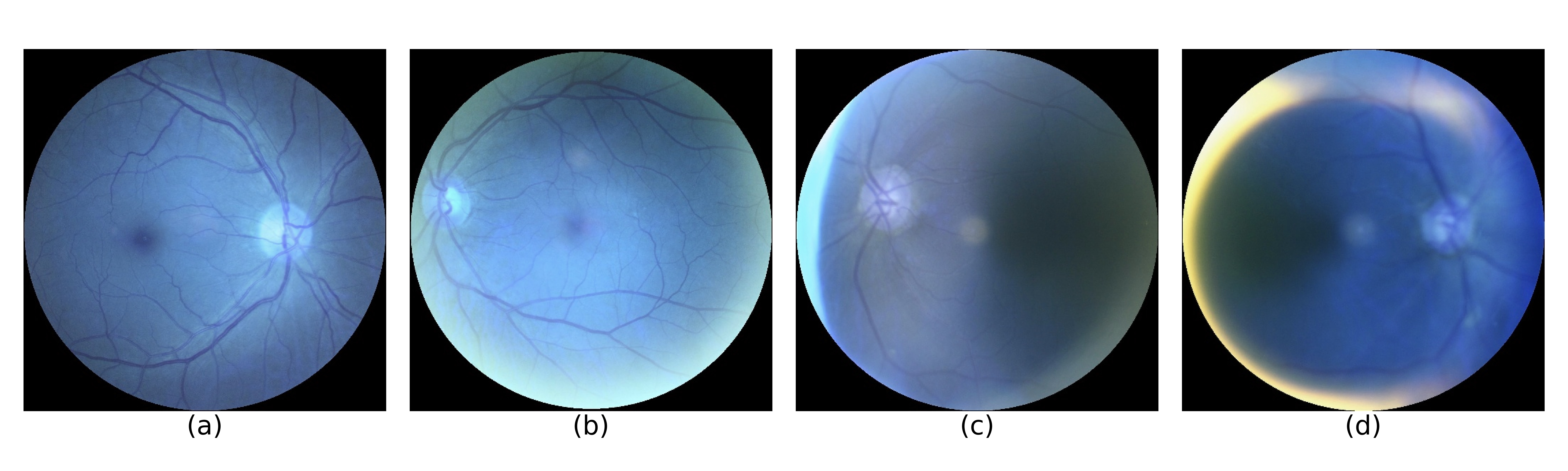}
  \caption{Visualization of sample images in the ODIR dataset. (a) and (b) are normal fundus images and (c) and (d) are cataractous fundus images.}
  \label{fig:fig1}
\end{figure}

\subsection{Dataset Preprocessing}
The ODIR dataset consisted of 594 images labeled as cataract and 5,675 images labeled as normal. To avoid an imbalanced dataset, 594 images labeled as normal were randomly selected. Hence the dataset then consisted of 1,188 images in total. Subsequently, image augmentation was applied to increase the size of the dataset to avoid overfitting and aid in model’s generalization and also to mimic real world scenarios. Two augmentations were applied to the original dataset, 30 degree rotation and -30 degree rotation. After augmentation, the dataset consisted of 3,564 images. The dataset was then split into training and testing sets in a 70:30 ratio.

\begin{figure}[H]
  \centering
  \includegraphics[width=0.68\textwidth]{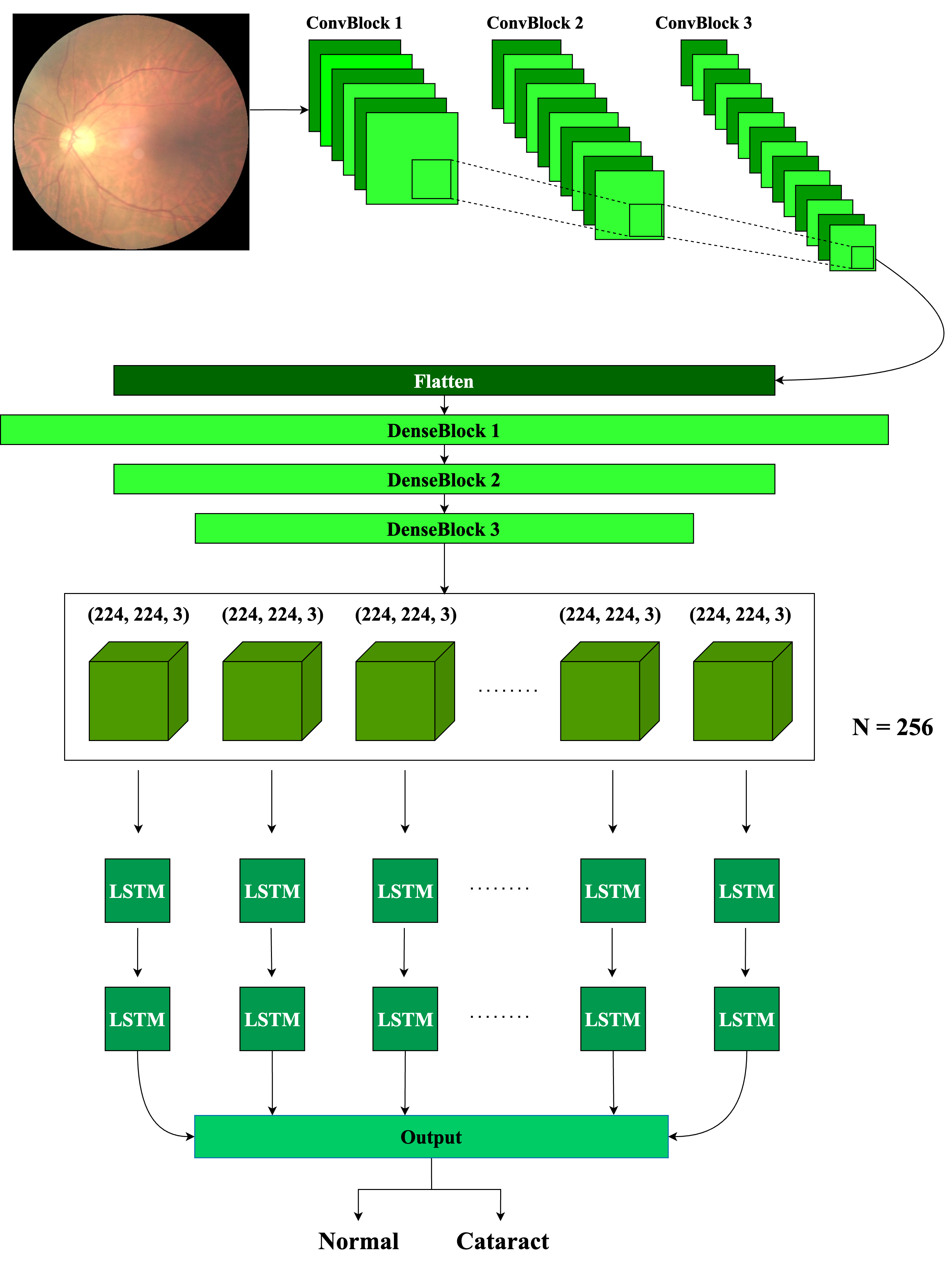}
  \caption{Architecture diagram of the proposed CNN-LSTM Model.}
  \label{fig:fig1}
\end{figure}

\subsection{Convolutional Neural Network}
Convolutional Neural Networks \cite{o2015introduction} are quite similar to typical neural networks, but unlike multilayer perceptrons, ConvNets assume the input is a two-dimensional image, which aids in learning intricate image characteristics and attributes. CNNs have various applications such as image classification, image segmentation, object identification, image creation, and so on, and ConvNets have shown to be particularly beneficial in medical imaging as a result of these applications. A Convolutional Neural Network comprises three main types of layers, convolutional, pooling, and Dense (fully connected) layers.
\\\\

CNN's foremost and most crucial layer is the convolutional layer. Convolution is a commonly used image processing method that has been used for a long time in which two image inputs are multiplied together to produce some output. The convolutional layer performs the same operation, multiplying the input picture by a collection of filters or kernels that are learned and altered throughout the training process, resulting in a smaller convoluted image that is forwarded to the next layer. The convolution operation follows this formula, where I is the input image and K is the kernel.

\begin{equation}
    S(i, j) = (I \cdot K)(i, j) = \sum_m \sum_n I(m, n)  K(i-m, j-n)
\end{equation}

Second, in the ConvNet, the Pooling layer is employed in between consecutive convolutional layers. The pooling layer's main function is to gradually lower the dimension of the image, pass just the significant information to the next layer, and reduce the amount of network parameters. The pooling layers also aid in the control of overfitting. Pooling layers are classified into three types: maximum pooling, average pooling, and global pooling. Max pooling determines the highest possible value for each patch in the feature map. Calculating the average for each patch of the feature map is what average pooling entails. Finally, global pooling reduces the feature map to a single value. Max pooling is the most commonly used pooling technique because it extracts the most important features like the bright regions and the edges. Finally, towards the network's end, the fully connected or dense layer is typically used. The neurons in the dense layer are linked to every neuron in the previous layer. The dense layer performs matrix-vector multiplication in the background.

\subsection{Long Short Term Memory}
Long Short Term Memory (LSTMs) \cite{greff2016lstm} are a kind of Recurrent Neural Network (RNN) (RNNs) \cite{sherstinsky2020fundamentals}. Unlike RNNs, LSTMs do remarkably well on long-term data. RNNs include a single tanh layer, however LSTMs have a unique structure with four modules that helps to prevent the vanishing gradient problem that occurs in RNNs. The key feature of LSTM is the cell state, which may add or delete information based on the output of structures known as gates. LSTMs have three gates: a forget gate, an input gate, and an output gate, where xt represents the current input, Ct and Ct 1 represent the new and previous cell states, respectively, and ht and ht 1 represent the present and previous outputs.
\\\\
The sigmoid layer in the "Forget gate" decides whether to keep or discard the information gained in the previous step. The forget gate follows the formula represented by Equation 2. The sigmoid layer produces a 0 or a 1. Output 0 means "delete the previous information," whereas output 1 means "keep the previous information."

\begin{equation}
    f_{t} = \sigma (W_{f} \cdot [h_{t-1}, x_{t}] + b_{f})
\end{equation}

The LSTM then chooses what additional information should be added to the cell state. The "Input gate" performs this action. It is composed of two layers: sigmoid and tanh. The sigmoid determines which values must be updated, and the tanh generates a list of candidate values that may be added to the cell state. Finally, the old state must be updated by multiplying it by ft (values to forget) and adding it to the new candidate values. The "Input gate" is represented by Equations 3, 4, and 5.

\begin{equation}
    i_{t} = \sigma (W_{i} \cdot [h_{t-1}, x_{t}] + b_{i})
\end{equation}

\begin{equation}
    \widetilde{C_{t}} = tanh(W_{C} \cdot [h_{t-1}, x_{t}] + b_{c})
\end{equation}

\begin{equation}
    C_{t} = f_{t} \cdot C_{t-1} + i_{t} \cdot  \widetilde{C_{t}}
\end{equation}

Finally, the "Output gate" determines the LSTM's output values. To begin, run the output through a sigmoid function. The cell state is then sent via a tanh layer so that the values range between -1 and 1. Finally, the sigmoid and tanh outputs are multiplied to get the LSTM output. The operation of the "Output gate" is represented by Equations 6 and 7.

\begin{equation}
    o_{t} = \sigma (W_{o} \cdot [h_{t-1}, x_{t}] + b_{o})
\end{equation}

\begin{equation}
    h_{t} = o_{t} \cdot tanh(C_{t})
\end{equation}

\subsection{Proposed Architecture}
This paper offers a unique CNN-LSTM combination model for classifying ocular fundus images as normal or cataractous. The network consists of 16 layers: one input layer, three convolutional layers, three max pooling layers, three batch normalization layers, one dropout layer, one flatten layer, one fully connected dense layer, two LSTM layers, and one output dense layer with sigmoid activation function. The input layer takes images with the dimensions 224 × 224 x 3. Then it's linked to three further sets of three layers each. Each set consists of three layers: convolutional, max pooling, and batch normalization. The convolutional layer has a 3 x 3 kernel and a ReLU \cite{agarap2018deep} activation function. The pool size used by the max pooling layer is 2 × 2. The output of the three sets of layers is transmitted to a Dropout layer with a dropout rate of 0.2, dropping 20\% of the neurons at random, followed by a Flatten layer to reduce the two-dimensional representation to a one-dimensional vector. The 1D vector is then transmitted to a fully connected Dense layer with 256 units and a ReLU activation function, followed by two LSTM layers with 256 units each and a tanh activation function. Finally, the final LSTM layer's output is transferred to a Dense layer with sigmoid activation for classification. Figure 3 depicts the proposed CNN-LSTM network's design.

\begin{figure}[H]
  \centering
  \captionsetup{justification=centering,margin=1cm}
  \includegraphics[width=\textwidth]{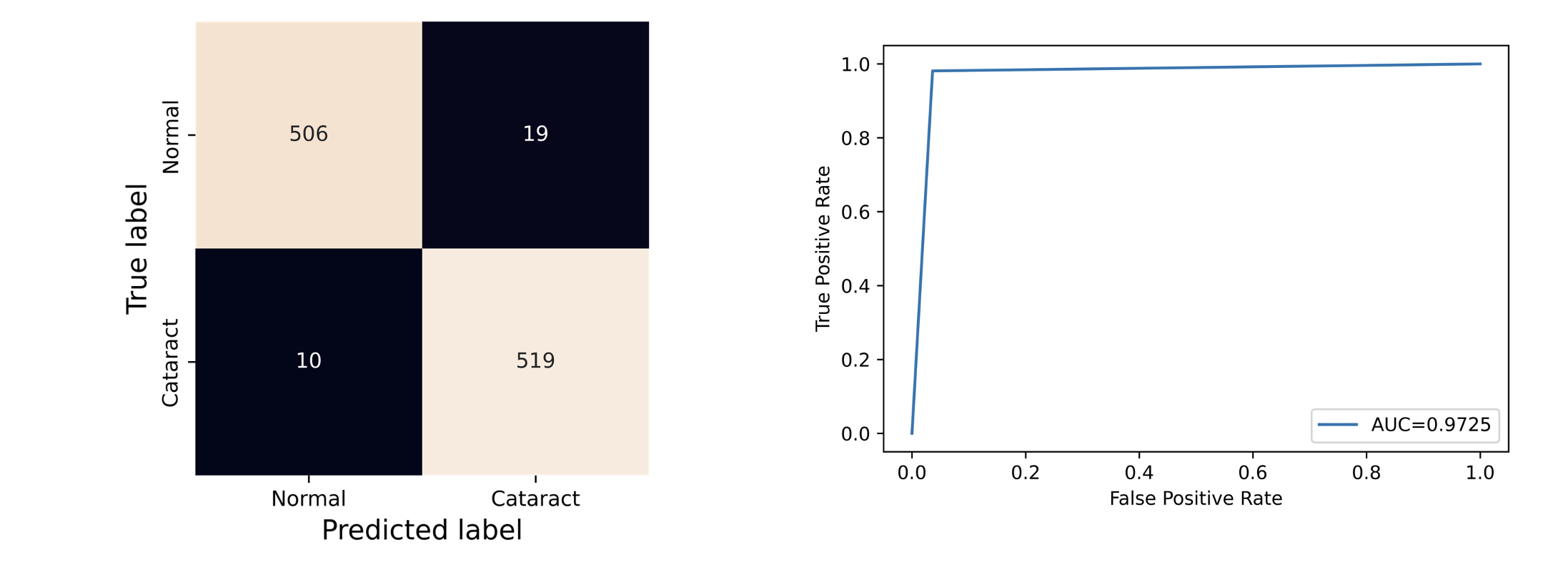}
  \caption{(a) Confusion matrix of the proposed model on the test set. (b) ROC (Receiver Operating Characteristic) Curve of the results on test set.}
  \label{fig:fig1}
\end{figure}

\section{Results and Discussion}

\subsection{Evaluation Metrics}
Performance metrics are used to evaluate the model's efficiency. These performance indicators show how well our model worked with the provided data. The proposed study evaluates CNN-LSTM architecture using the following six metrics: accuracy, precision, recall, sensitivity, specificity, and F1 score.

\subsubsection{Accuracy}
The accuracy of a classification algorithm is one technique to determine how well the algorithm correctly classifies a data point. 

\subsubsection{Precision}
Precision, also known as positive predictive value, is the accuracy of a model's positive prediction.

\subsubsection{Recall}
The recall of the model assesses its ability to recognize positive samples. The more positive samples identified, the larger the recall.

\subsubsection{Sensitivity}
Sensitivity is used to assess model performance since it shows how many positive instances the model accurately identified. It is same as Recall.

\subsubsection{Specificity}
The percentage of true negatives accurately recognized by the model is referred to as specificity.

\subsubsection{F1 Score}
The F1-score combines a classifier's precision and recall into a single metric by taking their harmonic mean.

\subsection{Results}
The proposed CNN-LSTM combination model achieved remarkable results on the testing set. The model achieved an accuracy, precision, recall, sensitivity, specificity, and F1 score of 97.53\%, 95.64\%, 99.62\%, 100\%, 98.48\%, and 97.59\%.

\section{Conclusion and Future Scope}
In this study, we explored and developed a combined CNN-LSTM model for classifying ocular fundus images as normal or cataractous. We trained the model on the renowned ODIR dataset. The model achieved inordinate results on both seen as well as unseen data. It achieved a training accuracy of 99.918\% and a testing accuracy of 97.24\%. Our proposed model fetched comparatively better results than the existing CNN-based as well as other pre-trained models utilized for ocular disease classification models while at the same time lowering time and space complexities.
\\\\
In the future, we plan on generalizing the model for multiple ocular diseases including myopia, glaucoma, etc, and obtain satisfactory results to actuate the system into a sizeable product.  As the world is moving towards complex ocular diseases, this system will be able to provide cheap and mobile ocular diagnostic solutions potentially reducing diagnostic time and enhancing treatment by providing deeper insights.

%Bibliography
\bibliographystyle{unsrt}
\bibliography{references}

\end{document}